\documentclass{article}

\usepackage[square,sort,comma,numbers]{natbib}
\usepackage[accepted]{icml2020}
\usepackage{url}            
\usepackage{booktabs}       
\usepackage{amsfonts}       
\usepackage{microtype}      
\usepackage{amssymb}
\usepackage{amsmath}
\usepackage{amsfonts}
\usepackage{amsthm}
\usepackage{algorithm}
\usepackage{algorithmic}
\usepackage{graphicx}
\usepackage{enumitem}
\usepackage{titlesec}
\newtheorem{problem}{Problem}
\newtheorem{theorem}{Theorem}

\newtheorem{definition}{Definition}
\usepackage{authblk}
\begin{document}
\twocolumn[
\icmltitle{Tunable Subnetwork Splitting for Model-parallelism of Neural Network Training}]
\begin{icmlauthorlist}
\icmlauthor{Junxiang Wang}{to}
\icmlauthor{Zheng Chai}{to}
\icmlauthor{Yue Cheng}{to}
\icmlauthor{Liang Zhao}{to}
\end{icmlauthorlist}
\icmlaffiliation{to}{George Mason University, United States}
\icmlcorrespondingauthor{Junxiang Wang}{jwang40@gmu.edu}
\printAffiliationsAndNotice{}
\begin{abstract}
Alternating minimization methods have recently been proposed as alternatives to the gradient descent for deep neural network optimization. Alternating minimization methods can typically decompose a deep neural network into layerwise subproblems, which can then be optimized in parallel.  Despite the significant parallelism, alternating minimization methods are rarely explored in training deep neural networks because of the severe accuracy degradation. In this paper, we analyze the reason and propose to achieve a compelling trade-off between parallelism and accuracy by a reformulation called Tunable Subnetwork Splitting Method (TSSM), which can tune the decomposition granularity of deep neural networks. Two methods gradient splitting Alternating Direction Method of Multipliers (gsADMM) and gradient splitting Alternating Minimization (gsAM) are proposed to solve the TSSM formulation. Experiments on five benchmark datasets show that our proposed TSSM can achieve significant speedup without observable loss of training accuracy. The code has been released at \url{https://github.com/xianggebenben/TSSM}.
\end{abstract} 
 \vspace{-1cm}
\section{Introduction}
\label{sec:introduction}
\vspace{-0.2cm}
 \indent Deep learning has gained unprecedented success during the last decade in a variety of applications such as healthcare and social media~\citep{miotto2018deep,zhao2015simnest}. Over recent years, the sizes of state-of-the-art deep learning models have increased significantly fast from millions of parameters to over billions \citep{he2016deep}. Therefore, efficient and scalable training of a deep neural network in time is crucial and has drawn much attention from the machine learning community. Traditional gradient-based methods such as Stochastic Gradient Descent (SGD) and Adaptive moment estimation (Adam) play a  dominant role in training neural networks. They 
train deep neural networks by the well-known backpropagation algorithm~\citep{leung1991complex}.  However, the backpropagation algorithm maybe computationally expensive when training very deep neural networks. This is because the backpropagation algorithm has poor concurrency and is difficult to be parallelized by layers, since the weights in one layer have to wait before all its dependent layers are updated, which is known as ``backward locking''~\citep{huo2018decoupled}. One famous recent work has proposed Decoupled Network Interface (DNI) to break backward locking, where a true gradient is replaced with an estimated synthetic gradient. However, extensive experiments show that the performance of DNI degraded seriously for very deep neural networks such as residual neural networks (ResNet) due to extra estimation errors~\citep{huo2018training,huo2018decoupled}.
 \\
\indent 
To address this challenge, some researchers have explored the possibility of alternating minimization methods as alternatives. These methods first decompose a large-scale deep neural network training problem into layerwise subproblems, each of which can be solved separately by efficient algorithms or closed-form solutions. Despite the good potential, the main challenge of alternating minimization methods consists in the nontrivial accuracy loss in training deep neural networkss during the decomposition process, because it is required to introduce extra constraints as well as possible relaxations. For example, our experiments show that the accuracy of ADMM was degraded to about $70\%$ when training a 10-layer feed-forward neural network, while SGD reached over $95\%$. The investigation results demonstrate that the ``decomposition'' of a deep neural network  is a double-edged sword that can improve the parallelism\footnote{The number of computing devices, e.g., GPUs, which can be used to train a model in parallel.}. while lowering down accuracy. The deeper a neural network is, the more profound such a trend is.\\
\indent In sum, gradient-based methods can achieve good accuracy but low parallelism, while, on the other hand, alternating minimization-based methods show good parallelism but low accuracy. 
Given that neither of the above two is robust enough for modern large-scale deep neural networks, in this paper, \emph{we achieve a compelling trade-off between them, with a novel reformulation of neural network training framework named Tunable Subnetwork Splitting Method (TSSM) to control the decomposition granularity of deep neural networks  and the degree of relaxations}. 
Under this reformulation, we further propose two methods to solve it, which are gradient splitting Alternating Direction Method of Multipliers (gsADMM), and gradient splitting Alternating Minimization (gsAM). 
\section{Tunable Subnetwork Splitting Method (TSSM)}
\label{sec:tunable subnetwork splitting}
\indent In this section, we present our novel problem formulation and two model-parallelism based training algorithms. Specifically, Section \ref{sec:problem formulation} introduces the deep neural network training problem via tunable subnetwork splitting. Sections \ref{sec:gsADMM} and \ref{sec:gsAM} present the gsADMM and gsAM algorithms, respectively.
\subsection{Problem Formulation}
\label{sec:problem formulation}
\begin{table}
\scriptsize
 \centering
 \caption{ Notations Used in This Paper}
 \begin{tabular}{cc}
 \hline
 Notations&Descriptions\\ \hline
 $n$& Number of subnetworks.\\
 $M$& Number of training samples.\\
 $y$& the predefined label vector.\\ 
 $W_l$& The weight of the $l$-th subnetwork.\\
 $p_l$& The input of the $l$-th subnetwork.\\
 $f_l (p_l)$& The architecture of the $l$-th subnetwork.\\
 $q_l$& The output of the $l$-th subnetwork.\\
 $R (W_n,p_n;y)$& The loss function in the $n$-th subnetwork.\\
\hline
 \end{tabular}
 \label{tab:notation}
 \end{table}
 \indent All notations used in this paper are outlined in Table \ref{tab:notation}.
We assume that a typical deep neural network consists of $n$ subnetworks, $p_l$, $q_l$ and $W_l$ denote the input, output and weights of the $l$-th subnetwork, respectively. $p_1$ is the input of both the first subnetwork and the whole neural network, $y$ is a predefined training label vector, and  $M$ is the number of the training set. The deep neural network training problem can be formulated as follows:
\begin{problem}
\label{prob:problem 1}
\small
\begin{align*}
     & \min\nolimits_{W_l,p_l} R (W_n,p_n;y) \\
     &s.t.\ q_l=f_l (W_l,p_l), \ p_{l+1}=q_l (l=1,\cdots,n-1)
\end{align*}
\normalsize
\end{problem}

where $f_l$ is the architecture of the $l$-th subnetwork, and $R (\bullet)$ is a continuous loss function on the $n$-th subnetwork. 
The constraint $q_l=f_l (W_l,p_l) (l=1,\cdots,n-1)$ is difficult to handle here, this is because the architecture of a neural network $f_l$ is highly nonlinear. To handle this challenge,  we relax Problem \ref{prob:problem 1} to Problem \ref{prob:problem 2} by imposing penalties on the objective as follows:
\begin{problem}
\label{prob:problem 2}
\small
\begin{align*}
    &\min\nolimits_{W_l,p_l} R (W_n,p_n;y)+\sum\nolimits_{l=1}^{n-1} \Omega (W_l,p_l,q_l) \ 
     \\&s.t. \ p_{l+1}=q_l (l=1,\cdots,n-1)
\end{align*}
\normalsize
\end{problem}
where the penalty is $\Omega (W_l,p_l,q_l)= (\alpha/2M)\Vert q_l-f_l (W_l,p_l)\Vert^2_2$ and $\alpha>0$ is a tuning parameter. It is easy to verify that $R (W_n,f_{n-1} (W_{n-1},\cdots,f_1 (W_1,p_1),\cdots,);y)$ is the loss function of Problem \ref{prob:problem 1} by replacing the loss with constraints $q_l=f_l (W_l,p_l)$ and $p_{l+1}=q_l$ recursively, and $R (W_n,p_n;y)$ is the loss function of Problem \ref{prob:problem 2}. 

Here we demonstrate how the relaxation influences the model accuracy by Theorem \ref{thero:approximation error}. Denote $r_l=q_l-f_l (W_l,p_l)$ as the relaxed term in the $l$-th subnetwork, the following theorem demonstrates the approximation error via $r_l$.
\begin{theorem}[Approximation Error]
\label{thero:approximation error} We define $\Vert R (W_n,f_{n-1} (W_{n-1},\cdots,f_1 (W_1,p_1),\cdots,);y)\!-\!R (W_n,p_n;y)\Vert$ as the approximation error.
If $f_l (W_l,p_l) (l=1,\cdots,n-1)$ and $R (W_n,p_n;y)$ are Lipschitz continuous with respect to $p_l$ and $p_n$ with coefficient $H_l (W_l)>0$ and $H_n (W_n)>0$, respectively, then for any $ (W_1,\cdots,W_n)$ we have
\small
\begin{align}
\nonumber&\Vert R (W_n,f_{n-1} (W_{n-1},\cdots,f_1 (W_1,p_1),\cdots,);y)\!-\!R (W_n,p_n;y)\Vert\\&\leq H_n (W_n)\sum\nolimits_{l=1}^{n-1} (\Vert r_l\Vert \prod\nolimits_{j=l\!+\!1}^{n\!-\!1} H_{j} (W_{j}))
    \label{eq:approximation error}
\end{align}
\end{theorem}
\normalsize
\indent The definition of Lipschitz continuity is given in the Appendix. The assumption in Theorem \ref{thero:approximation error} is  mild to satisfy: most common networks such as fully-connected neural networks and Convectional Neural Networks (CNN) are Lipschitz continuous \citep{virmaux2018Lipschitz}; common loss functions such as the cross-entropy loss and the least square loss are Lipschitz continuous as well \citep{wang2019admm}.
\indent Theorem \ref{thero:approximation error} states that the gap between the loss functions of Problems \ref{prob:problem 1} and \ref{prob:problem 2} is upper bounded by the linear combination of $r_i$. The larger $n$ is (i.e. The more subnetworks we split a neural network into), the larger upper bound of the approximation error is (i.e., the worse training performance a neural network has), and  the finer-gained a deep neural networks is parallelized during training. In other words, this theorem indicates that there is a trade-off between the degree of parallelism and performance.
This also explains why existing ADMM methods drastically degrade the accuracy in the deep neural network problems: they decompose a deep neural network into layer-wise components, where all layers involve approximation errors. And hence the errors drastically increase while deep neural networks is deeper. And our proposed formulation aims to control the approximation error and degree of relaxation via $n$ toward model parallelism without too much loss of performance. In the next two sections, we present two novel model-parallelism based algorithms to solve this formulation.
\subsection{The gsADMM Algorithm}
\label{sec:gsADMM}
\begin{algorithm}[H] 
\scriptsize
\caption{the gsADMM Algorithm} 
\begin{algorithmic}
\label{algo: gsADMM}
\REQUIRE $y$, $p_1$, $\rho$, $\alpha$. 
\ENSURE $\textbf{W},\textbf{p},\textbf{q}$. 
\STATE Initialize $k=0$.
\WHILE{$\textbf{W}^k,\textbf{p}^{k},\textbf{q}^{k}$ do not converge}
\STATE Choose the sample index set $s\subset [1,2,3,\cdots,M]$ of size $b$ uniformly at random and generate $ (p^k_{l,s},q^k_{l,s}) (l=1,\cdots,n-1)$ and $ (p^k_{n,s},y_s)$.
\STATE $W^{k+1}_l$ is updated by SGD or Adam in parallel.
\STATE Update $p_l^{k+1}$ by Equation \eqref{eq:update p} in parallel.
\STATE Update $q_l^{k+1}$ by Equation \eqref{eq:update q} in parallel.
\STATE Update $u_l^{k+1}$ by Equation \eqref{eq:update u} in parallel.
\STATE $k\leftarrow k+1$.
\ENDWHILE
\STATE Output $\textbf{W},\textbf{p},\textbf{q}$.
\end{algorithmic}
\end{algorithm}

\indent Now we follow the standard rountine of ADMM to solve Problem \ref{prob:problem 2}.
The augmented Lagrangian function is mathematically formulated as $L_\rho (\textbf{W},\textbf{p},\textbf{q},\textbf{u})=R (W_n,p_n;y)+\sum\nolimits_{l=1}^{n-1} \Omega (W_l,p_l,q_l)+\sum\nolimits_{l=1}^{n-1} u_l^T (p_{l+1}-q_l)+ (\rho/2)\sum\nolimits_{l=1}^{n-1}\Vert p_{l+1}-q_l\Vert^2_2$ where $\rho>0$ is a penalty parameter and $u_{l}$ is a dual variable. To simplify the notations, we denote
$\textbf{p}=\{p_l\}_{l=1}^n$,$\textbf{W}=\{W_l\}_{l=1}^n$, $\textbf{q}=\{q_l\}_{l=1}^{n-1}$, $\textbf{u}=\{u_l\}_{l=1}^{n-1}$,
$\textbf{W}^{k+1}=\{W_l^{k+1}\}_{l=1}^L$, $\textbf{p}^{k+1}=\{p_l^{k+1}\}_{l=1}^L$,  $\textbf{q}^{k+1}=\{q_l^{k+1}\}_{l=1}^L$, $\textbf{u}^{k+1}=\{u_l^{k+1}\}_{l=1}^L$, $\textbf{p}^{k+1}_s=\{p_{l,s}^{k+1}\}_{l=1}^L$,  $\textbf{q}^{k+1}_s=\{q_{l,s}^{k+1}\}_{l=1}^L$. The core idea of the gsADMM algorithm is to update one variable alternately while fixing others, which is  shown in Algorithm \ref{algo: gsADMM}. Specifically, 
in Line 5, a sample index set $s$ of size $b$ is chosen randomly from the training set of size $M$ , in order to generate the batch training set for the $l$-th subnetwork, which are $ (p^k_{l,s},q^k_{l,s}) (l=1,\cdots,n-1)$ and $ (p^k_{n,s},y_s)$. Lines 7, 9, 10 and 11 update $W_l$, $p_l$, $q_l$, and $u_l$, respectively. All subproblems are discussed as follows: \\
\textbf{1. Update $\textbf{W}$}\\
\indent This subproblem can be updated via either SGD or Adam. Take the SGD as an example, the solution is shown as follows:
\small
\begin{align}
&W_l^{k+1}\leftarrow W_l^{k}- \nabla_{W_l} \Omega (W^k_l,p^k_{l,s},q^k_{l,s})/\tau_1^{k+1} (l=1,\cdots,n-1)\label{eq:update W_l}
\\& W_n^{k+1}\leftarrow W_n^{k}- \nabla_{W_n} R (W^k_n,p^k_{n,s};y_s)/\tau_1^{k+1}\label{eq:update W_n}
\end{align}
\normalsize
where $1/\tau_1^{k+1}>0$ is a learning rate for SGD on the $ (k+1)$-th epoch.\\
\textbf{2. Update $\textbf{p}$}\\
The solution is shown as follows:
\small
\begin{align}
    \textbf{p}^{k+1}\leftarrow \textbf{p}^{k}-\nabla_{\textbf{p}} L_\rho (\textbf{W}^{k+1},\textbf{p}^{k},\textbf{q}^{k},\textbf{u}^k)/\tau_2^{k+1} \label{eq:update p}
\end{align}
\normalsize
where $1/\tau^{k+1}_2>0$ is a learning rate.\\
\textbf{3. Update $\textbf{q}$}\\
This subproblem has a closed-form solution, which is shown as follows:
\small
\begin{align}
    q^{k+1}_l\leftarrow (\alpha f_l (W^{k+1}_l,p^{k+1}_l)+\rho M p^{k+1}_{l+1}+M u^k_l)/ (\rho M+\alpha)
    \label{eq:update q}
\end{align}
\normalsize
\textbf{4. Update $\textbf{u}$}\\
\indent The dual variable $\textbf{u}$ is updated as follows:
\small
\begin{align}
    u^{k+1}_l\leftarrow u^k_l+\rho (p^{k+1}_{l+1}-q^{k+1}_l)
    \label{eq:update u}
\end{align}
\normalsize
\subsection{The gsAM Algorithm}
\label{sec:gsAM}
\begin{algorithm}[H]

\scriptsize
\caption{the gsAM Algorithm} 
\begin{algorithmic}
\label{algo: gsAM}
\REQUIRE $y$, $p_1$, $\rho$, $\alpha$. 
\ENSURE $\textbf{W},\textbf{p},\textbf{q}$. 
\STATE Initialize $k=0$.
\WHILE{$\textbf{W}^k,\textbf{p}^{k}$ do not converge}
\STATE Choose the sample index set $s\subset [1,2,3,\cdots,M]$ of size $b$ uniformly at random and generate $ (p^k_{l,s},p^k_{{l+1},s}) (l=1,\cdots,n-1)$ and $ (p^k_{n,s},y_s)$.
\STATE $W^{k+1}_l$ is updated by SGD or Adam in parallel.
\FOR{l=2 to n}
\STATE Update $p_l^{k+1}$ by Equations \eqref{eq:update p_l} and \eqref{eq:update p_n}. 
\ENDFOR
\STATE $k\leftarrow k+1$.
\ENDWHILE
\STATE Output $\textbf{W},\textbf{p},\textbf{q}$.
\end{algorithmic}
\end{algorithm}
\indent The gsADMM algorithm can realize model parallelism while it may consume a lot of memory when training a large-scale deep neural network because of more extra variables involved. For example, for the deep Residual Network (ResNet) architecture, the dimension of $p_l,q_l$ and $u_l$ maybe as large as $50000\times 64\times 7\times 7$. To handle this challenge, we propose a variant of gsADMM called gsAM via reducing the number of auxiliary variables. To achieve this, we transform Problem \ref{prob:problem 2} equivalently to 
Problem \ref{prob:problem 3} via reducing the constraint $p_{l+1}=q_l$ as follows:
\begin{problem}
\label{prob:problem 3}
\small
\begin{align*}
    \min\nolimits_{W_l,p_l} F (\textbf{W},\textbf{p})= R (W_n,p_n;y)+\sum\nolimits_{l=1}^{n-1} \Omega (W_l,p_l,p_{l+1})
\end{align*}
\end{problem}
\normalsize
\indent Next we apply the gsAM algorithm to solve Problem \ref{prob:problem 3}, as shown in Algorithm \ref{algo: gsAM}. Specifically, 
Line 5 generates the batch training set for the $l$-th subnetwork, which are $ (p^k_{l,s},p^k_{l+1,s}) (l=1,\cdots,n-1)$ and $ (p^k_{n,s},y_s)$, and we assume that $F (\textbf{W}^k,\textbf{p}^k)=\mathbb{E} (F (\textbf{W}^k,\textbf{p}_s^k))$.  Lines 7, and 9 update $W_l$ and $p_l$ respectively. All subproblems are  discussed  as follows:\\
 \textbf{1. Update $\textbf{W}$}\\
Taking SGD as an example, the solution is shown as follows, where $1/\tau^{k+1}_1>0$ is a learning rate:
\small
\begin{align*}
    &W_l^{k+1}\leftarrow W_l^{k}- \nabla_{W_l} \Omega (W^k_l,p^k_{l,s},p^k_{{l+1},s})/\tau_1^{k+1} (l=1,\cdots,n-1)\\& W_n^{k+1}\leftarrow W_n^{k}- \nabla_{W_n} R (W^k_n,p^k_{n,s};y_s)/\tau_1^{k+1}
\end{align*}
\normalsize
\textbf{2. Update $\textbf{p}$}\\
\indent This can be solved via gradient descent as  follows, where $1/\tau^{k+1}_2>0$ is a learning rate:
\small
\begin{align}
&p^{k\!+\!1}_l\!\leftarrow\! p^{k}_l\!-\! (\nabla_{p_l}\Omega (W^{k\!+\!1}_{l\!-\!1},p^{k\!+\!1}_{l\!-\!1},p^k_{l})\!+\! \nabla_{p_l}\Omega (W^{k\!+\!1}_l\!,\!p^k_l\!,\!p^{k}_{l\!+\!1}))/\tau_2^{k\!+\!1} \label{eq:update p_l}\\
   &p^{k\!+\!1}_n\leftarrow p^{k}_n\!-\! (\nabla_{p_n}\Omega (W^{k\!+\!1}_{n\!-\!1},p^{k\!+\!1}_{n\!-\!1},p^k_{n})\!+\! \nabla_{p_n}R (W^{k+1}_n,p^k_n;y))/\tau_2^{k\!+\!1} \label{eq:update p_n}
\end{align}
\normalsize
\section{Experiment}
\label{sec:experiment}
In this section, we evaluate TSSM using five benchmark datasets. Effectiveness and speedup are compared with state-of-the-art methods on different neural network architectures. All experiments were conducted on a 64-bit Ubuntu 16.04 LTS server with Intel (R) Xeon CPU and GTX1080Ti GPU. 
\subsection{Experimental Settings}
\indent In this experiment, five benchmark datasets, MNIST~\citep{lecun1998gradient}, Fashion MNIST~\citep{xiao2017fashion}, kMNIST~\citep{clanuwat2018deep},  CIFAR10~\citep{krizhevsky2009learning} and CIFAR100~\citep{krizhevsky2009learning} are used for comparison.\\
\indent SGD~\citep{bottou2010large}, Adam~\citep{kingma2014adam} and the deep learning Alternating Direction Method of Multipliers (dlADMM)~\citep{wang2019admm} are state-of-the-art comparison methods. All hyperparameters were chosen by maximizing the accuracy of the training datasets. The learning rate of  SGD and Adam were set to 0.01 and 0.001, respectively. dlADMM's $\rho$ was set to $10^{-6}$.
Due to space limit, the details of all datasets and comparison methods can be found in the Appendix.
\subsection{Experimental Results}
\subsubsection{Performance}
\begin{figure}
  \centering
\begin{minipage}
{0.49\linewidth}
\centerline{\includegraphics[width=\columnwidth]{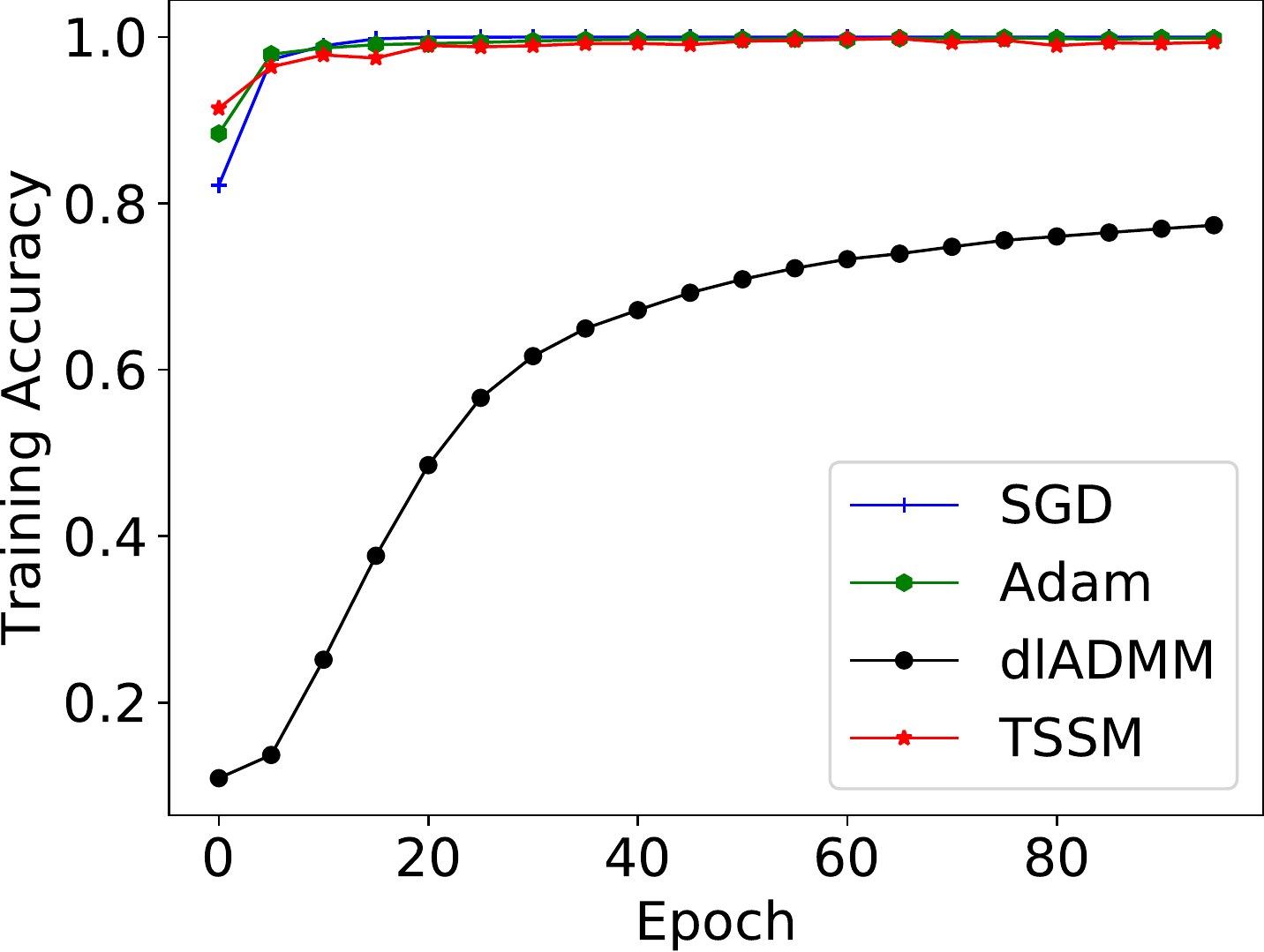}}
\centerline{ (a). MNIST}
\end{minipage}
\hfill
\begin{minipage}
{0.49\linewidth}
\centerline{\includegraphics[width=\columnwidth]{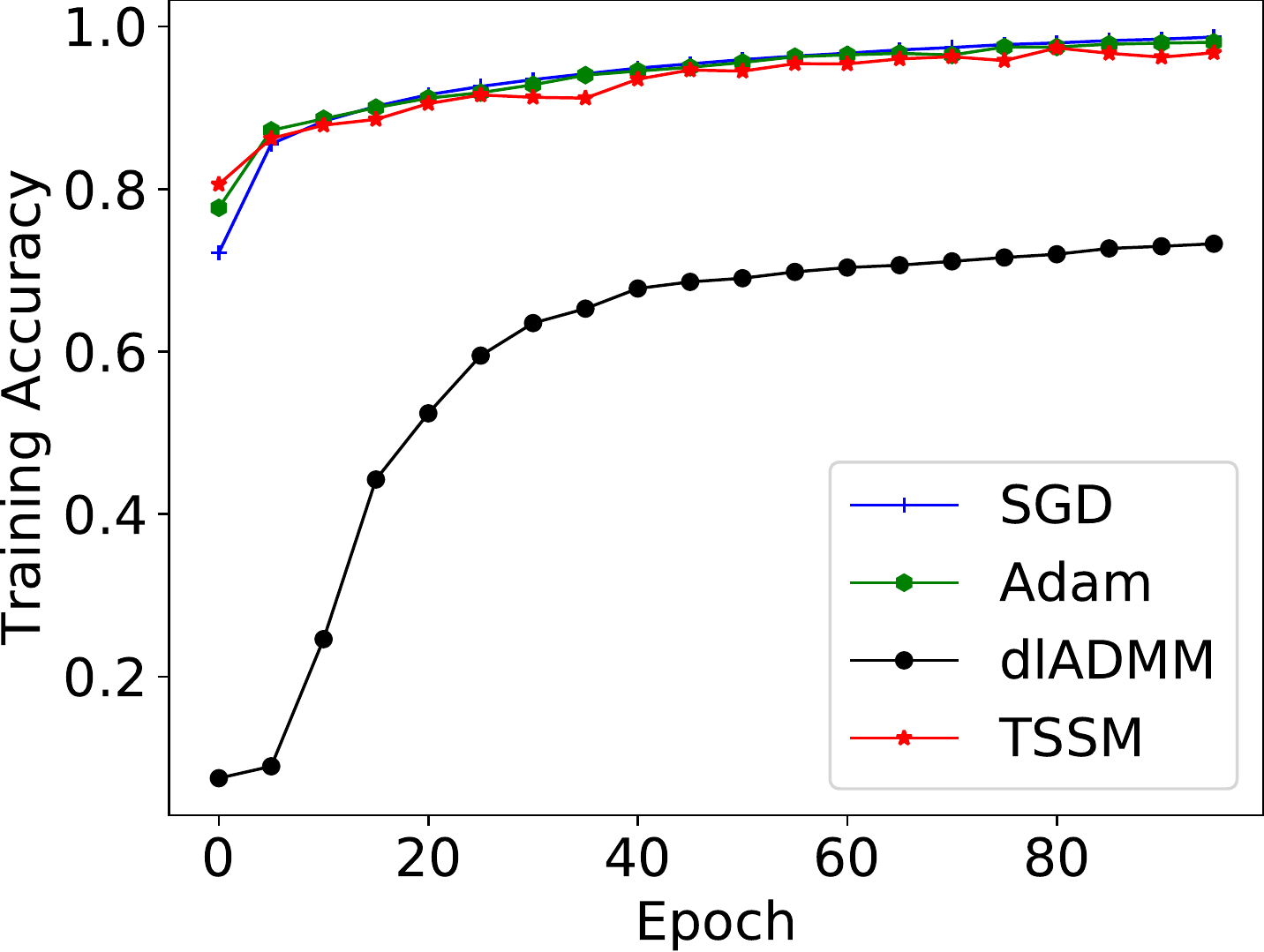}}
\centerline{ (b).  Fashion MNIST}
\end{minipage}
\hfill
\begin{minipage}
{0.49\linewidth}
\centerline{\includegraphics[width=\columnwidth]{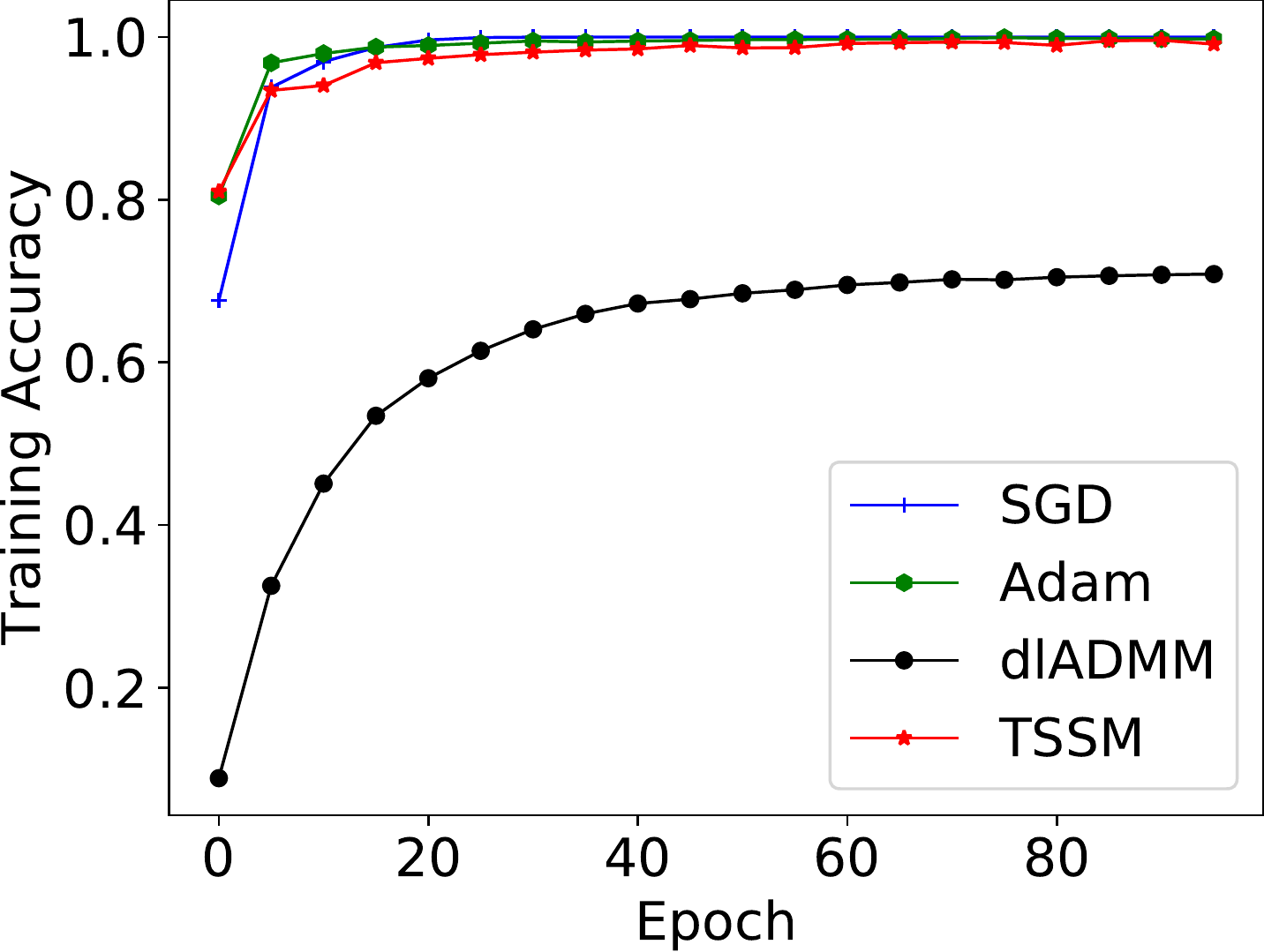}}
\centerline{ (c). kMNIST}
\end{minipage}
\hfill
\begin{minipage}
{0.49\linewidth}
\centerline{\includegraphics[width=\columnwidth]{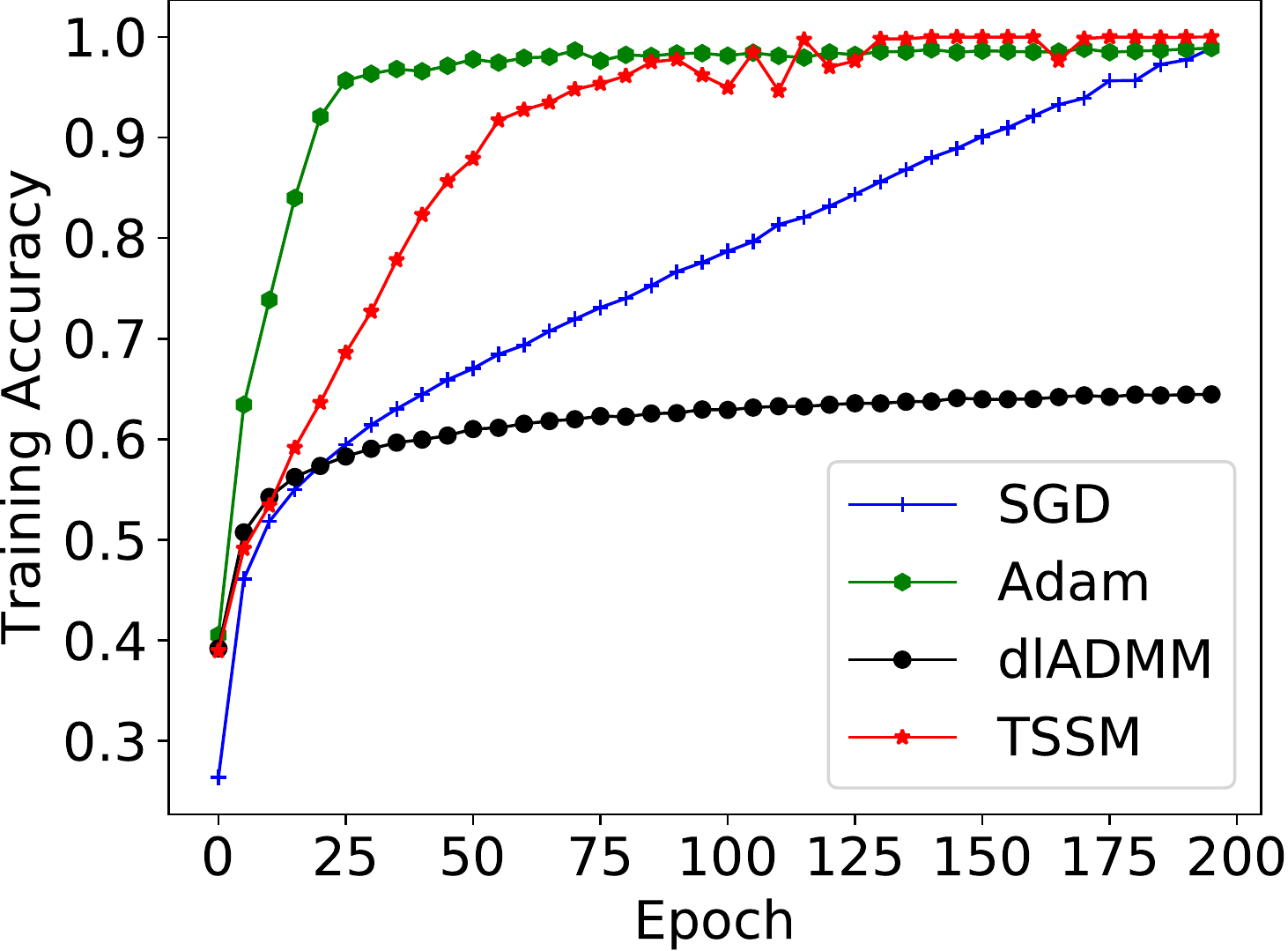}}
\centerline{ (d).  CIFAR10}
\end{minipage}
\hfill
\begin{minipage}
{0.49\linewidth}
\centerline{\includegraphics[width=\columnwidth]{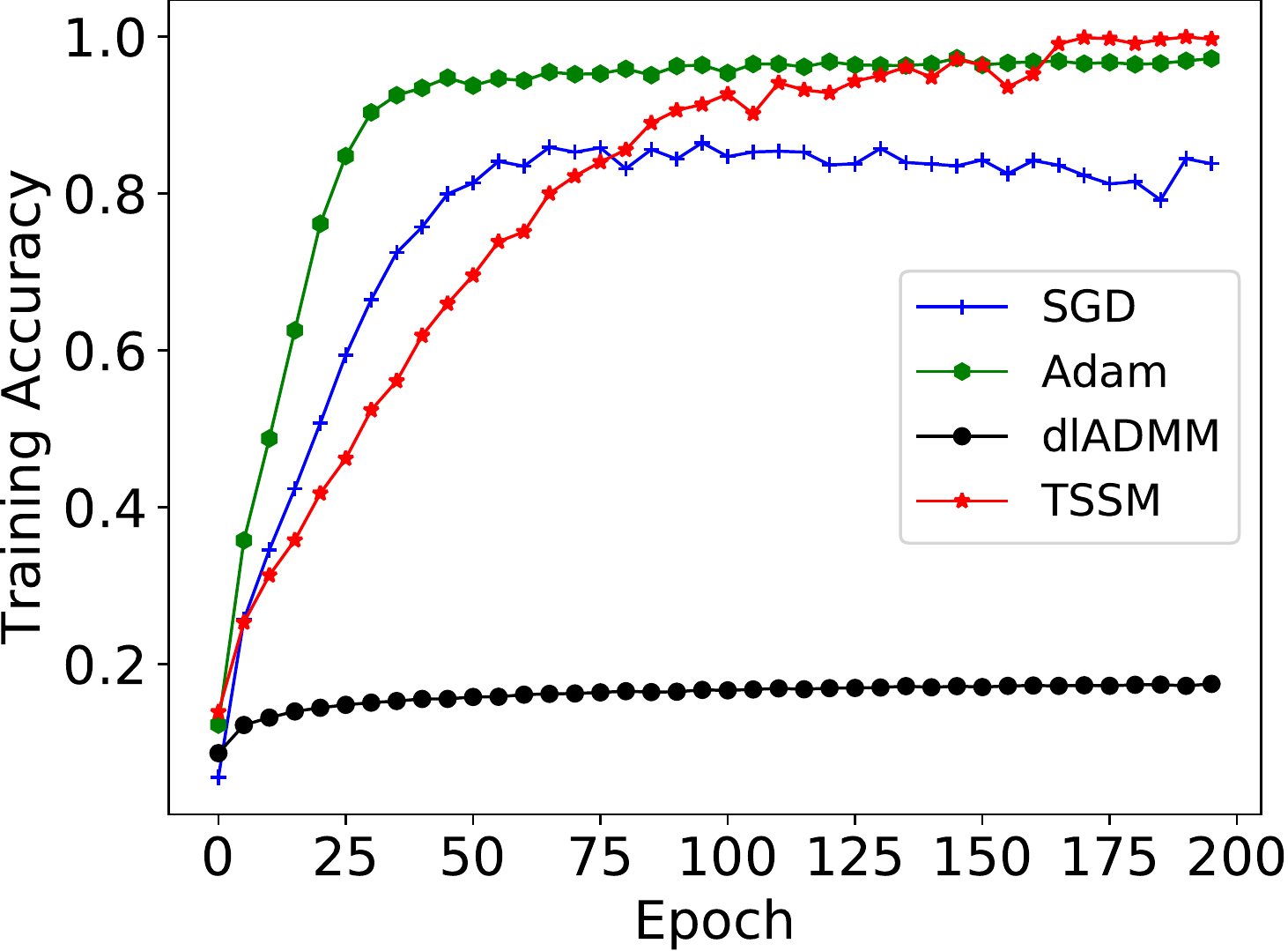}}
\centerline{ (e). CIFAR100}
\end{minipage}
  \caption{Training accuracy of all methods on five datasets. Our TSSM reaches the same training accuracy as SGD and Adam.}.
  \label{fig:performance}
  \vspace{-1cm}
\end{figure}
\indent We compare the performance of the TSSM with other methods. 
In this experiment, we split a network into two subnetworks. gsAM performs almost identical as gsADMM, and we only shows TSSM solved by gsADMM. \\
\indent We set up a feed-forward neural network which contained nine hidden layers with $512$ neurons each. The Rectified linear unit (ReLU) was used for the activation function. The loss function was set as the cross-entropy loss. $\alpha$ and $\rho$ were set to $1$. $\tau_1$ and $\tau_2$ were both set to $100$.
Moreover, we set up another Convolution Neural Network (CNN), which was tested on CIFAR10 and CIFAR100 datasets. This CNN architecture has two alternating stages of $3\times3$
convolution filters and $2\times2$ max pooling with no stride, followed by two fully connected layer
of $512$ rectified linear hidden units (ReLU’s). Each CNN layer has 32 filters. The number of epoch for feed-forward neural network and CNN was set to 100 and 200, respectively. The batch size for all methods except dlADMM was set to $120$. This is because dlADMM is a full-batch method.\\
\indent As shown in Figure~\ref{fig:performance}, our proposed TSSM reaches almost $100\%$ training accuracy on every dataset, which is similar to SGD and Adam. dlADMM, however, performed the worst on all datasets. This can be explained by our Theorem~\ref{thero:approximation error} that, dlADMM over-relaxes the neural network problems, and its performance becomes worse as the neural network gets deeper (compared with its excellent performance of shallow networks shown in~\citep{wang2019admm}).
\subsubsection{Speedup}
\begin{figure}
  \centering
\begin{minipage}
{0.49\linewidth}
\centerline{\includegraphics[width=\columnwidth]{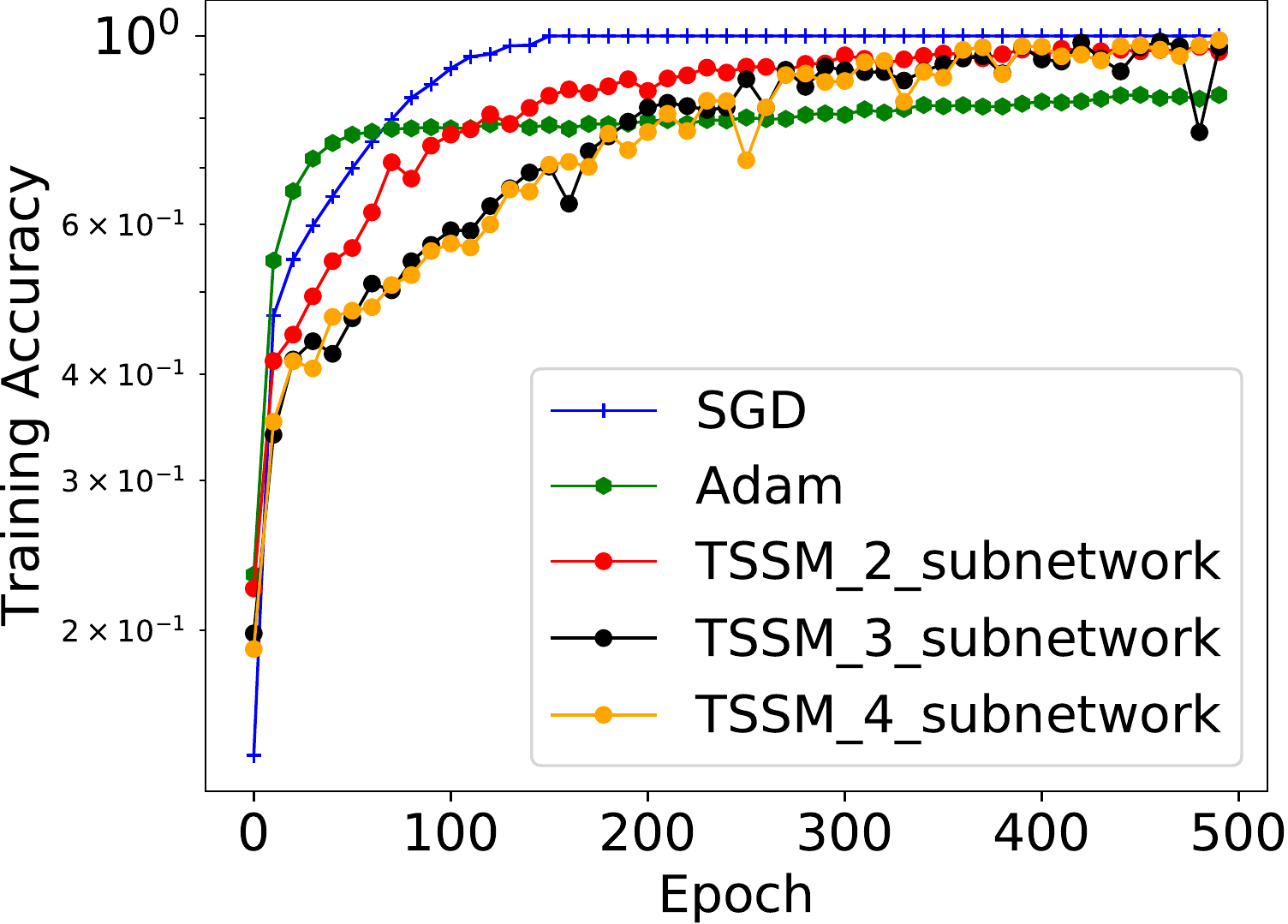}}
\centerline{ (a). Training accuracy.}
\end{minipage}
\hfill
\begin{minipage}
{0.49\linewidth}
\centerline{\includegraphics[width=\columnwidth]{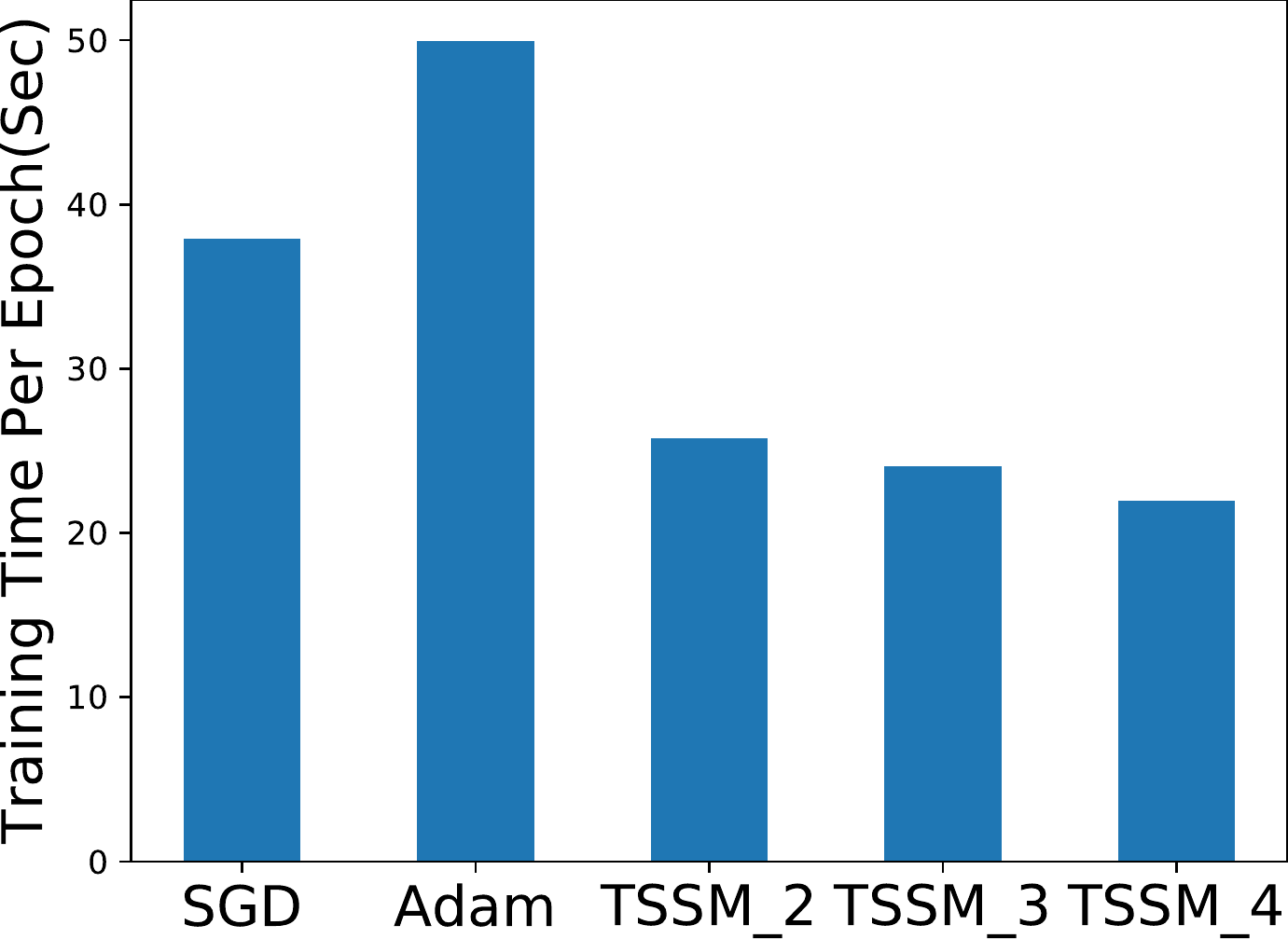}}
\centerline{ (b). Training time per epoch. }
\end{minipage}
  \caption{Training accuracy and training time on the ResNet-like architecture and CIFAR10 dataset: TSSM achieves the same performance as SGD yet more efficiently.}
  \label{fig:speedup}
\end{figure}
\indent Finally, we demonstrate how much speedup the TSSM can achieve for training large-scale neural networks. 
We test TSSM on a ResNet-like architecture. This network consists of  $9\times9$ convolution filters with stride of 3 and $3\times3$ max pooling with stride of 3 that are followed by 24 ResNet blocks, an average pooling and two fully connected layers of 512 rectified linear hidden units (ReLU’s). Each ResNet block is composed of two stages of $3\times3$ convolution filters.\\
\indent Figure~\ref{fig:speedup}(a) shows the training accuracy of SGD, Adam, and our proposed TSSM on the CIFAR10 dataset. dlADMM was not compared due to its inferior performance in the previous experiments. Specifically, we split this ResNet-like architecture into two, three, and four subnetworks.  In Figure~\ref{fig:speedup}(a), our TSSM reaches nearly $100\%$ training accuracy for all the subnetwork configurations that we tested, while Adam only achieves below $90\%$ training accuracy. Figure~\ref{fig:speedup}(b) compares the training time per epoch for all methods. TSSM$\_$2, TSSM$\_$3, TSSM$\_$4 represent 2, 3, and 4 subnetworks split by TSSM, respectively. The results were the average of 500 epochs. It takes at least 40 seconds for SGD and Adam to finish an epoch, while TSSM$\_$2 only takes 25 seconds per epoch. The training duration scales (sub-linearly) as TSSM is configured with more subnetworks.
TSSM can effectively reduce the training time by as much as $50\%$ 
without loss of training accuracy. This demonstrates TSSM's efficacy in both training accuracy and parallelism.
\vspace{-0.5cm}
\section{Conclusion}
\label{sec:conclusion}
\vspace{-0.3cm}
In this paper, we propose the TSSM and two mode-parallelism based algorithms gsADMM and gsAM to train deep neural networks. We split a neural network into independent subnetworks while controlling the loss of performance. Experiments on five benchmark datasets demonstrate the performance and speedup of the proposed TSSM.
\bibliographystyle{plain}
\bibliography{reference}
\newpage
\onecolumn
\large{Appendix}
\begin{appendix}
\textbf{The defintion of Lipschitz Continuity}\\\indent
The Lipschitz continuity is defined as follows:
\begin{definition} (Lipschitz Continuity)
A function $g (x)$ is Lipschitz continuous if there exists a constant $D>0$ such that $\forall x_1,x_2$, the following holds
\begin{align*}
    \Vert g (x_1)-g (x_2)\Vert\leq D\Vert x_1-x_2\Vert. 
\end{align*}
\end{definition}
\textbf{Proof of Theorem \ref{thero:approximation error}}
\begin{proof}
Because 
\begin{align}
    \nonumber&\Vert f_{l} (W_{l},f_{l-1} (\cdots,f_1 (W_1,p_1),\cdots,))-q_{l}\Vert\\\nonumber&= \Vert f_{l} (W_{l},f_{l-1} (\cdots,f_1 (W_1,p_1),\cdots,))-f_{l} (W_l,p_l)+f_{l} (W_l,p_l)-q_l \Vert\\\nonumber&\leq\Vert f_{l} (W_{l},f_{l-1} (\cdots,f_1 (W_1,p_1),\cdots,))-f_{l} (W_l,p_l)\Vert+\Vert f_{l} (W_l,p_l)-q_l \Vert \text{ (Triangle Inequality)}\\\nonumber&\leq H_l (W_l)\Vert f_{l-1} (\cdots,f_1 (W_1,p_1),\cdots,)-p_l \Vert+\Vert f_l (W_l,p_l)-q_{l}\Vert (\text{$f_l$ is Lipschitz continuous})\\\nonumber &=H_l (W_l)\Vert f_{l-1} (\cdots,f_1 (W_1,p_1),\cdots,)-q_{l-1} \Vert+\Vert f_l (W_l,p_l)-q_{l}\Vert (\text{$q_{l-1}=p_l$})\\&=H_l (W_l)\Vert f_{l-1} (\cdots,f_1 (W_1,p_1),\cdots,)-q_{l-1} \Vert+\Vert r_l\Vert (r_l=f_l (W_l,p_l)-q_l) \label{eq: recursion}
\end{align}
We apply recursively Equation \eqref{eq: recursion} to obtain
\begin{align*}
  &\Vert f_{n-1} (W_{n-1},f_{n-2} (\cdots,f_1 (W_1,p_1),\cdots,))-q_{n-1}\Vert\\&\leq  H_{n-1} (W_{n-1})\Vert f_{n-2} (\cdots,f_1 (W_1,p_1),\cdots,)-q_{n-2} \Vert+\Vert r_{n-1}\Vert\\&\leq H_{n-1} (W_{n-1})H_{n-2} (W_{n-2})\Vert f_{n-3} (\cdots,f_1 (W_1,p_1),\cdots,)-q_{n-3}\Vert+H_{n-1} (W_{n-1})\Vert r_{n-2}\Vert+\Vert r_{n-1}\Vert\\&\leq \cdots\\&\leq \sum\nolimits_{l=1}^{n-1} (\Vert r_l\Vert \prod\nolimits_{j=l+1}^{n-1} H_{j} (W_{j})) 
\end{align*}
Therefore
\begin{align*}
    &\Vert R (W_n,f_{n-1} (W_{n-1},\cdots,f_1 (W_1,p_1),\cdots,);y)-R (W_n,p_n;y)\Vert\\&\leq H_n (W_n)\Vert f_{n-1} (W_{n-1},\cdots,f_1 (W_1,p_1),\cdots,)-p_n\Vert\\&=H_n (W_n)\Vert f_{n-1} (W_{n-1},\cdots,f_1 (W_1,p_1),\cdots,)-q_{n-1}\Vert (\textbf{$q_{n-1}=p_{n}$})
\\&\leq H_n (W_n)\sum\nolimits_{l=1}^{n-1} (\Vert r_l\Vert \prod\nolimits_{j=l+1}^{n-1} H_{j} (W_{j}))
\end{align*}
\end{proof}
\section*{Experimental Details}
\subsection*{Dataset}
In this experiment, five benchmark datasets were used for comparison, all of which are provided by the Keras library \citep{chollet2015keras}.\\
1. MNIST \citep{lecun1998gradient}. The MNIST dataset has ten classes of handwritten-digit images, which was firstly introduced by Lecun et al. in 1998 \citep{lecun1998gradient}. It contains 55,000 training samples and 10,000 test samples with 196 features each. \\
2. Fashion MNIST \citep{xiao2017fashion}. The Fashion MNIST dataset has ten classes of assortment images on the website of Zalando, which is Europe’s largest online fashion platform \citep{xiao2017fashion}. The Fashion-MNIST dataset consists of 60,000 training samples and 10,000 test samples with 196 features each.\\
3. Kuzushiji-MNIST \citep{clanuwat2018deep}.  The Kuzushiji-MNIST dataset has ten classes, each of which is a character to represent each of the 10 rows of Hiragana. The Kuzushiji-MNIST dataset consists of 60,000 training samples and 10,000 test samples with 196 features each.\\
4. CIFAR10 \citep{krizhevsky2009learning}. CIFAR10 is a collection of color images with 10 different classes. The number of training data and test data are 60,000 and 10,000, respectively, with 768 features each.\\
5. CIFAR100 \citep{krizhevsky2009learning}. CIFAR100 is similar to CIFAR10 except that CIFAR100 has 100 classes. The number of training data and test data are 60,000 and 1,0000, respectively, with 768 features each.
\subsection*{Comparison Methods}
\indent The SGD, Adam, and the dlADMM are state-of-the-art comparison methods. All hyperparameters were chosen by maximizing the accuracy of the training dataset. \\
\textbf{1) Stochastic Gradient Descent (SGD)} \citep{bottou2010large}. SGD and its variants are the most popular deep learning optimizers, whose convergence has been studied extensively in the literature. The learning rate was set to $10^{-2}$.\\
\textbf{2) Adaptive momentum estimation (Adam)} \citep{kingma2014adam}. Adam is the most popular optimization method for deep learning models. It estimates the first and second momentum to correct the biased gradient and thus makes convergence fast. The learning rate was set to $10^{-3}$.\\
\textbf{3) deep learning Alternating Direction Method of Multipliers (dlADMM)} \citep{wang2019admm}. dlADMM is an improved version of ADMM for deep learning models. The parameter $\rho$ was set to $10^{-6}$.
\end{appendix}
\end{document}